\documentclass{article}
\usepackage{ijcai24}

\usepackage{times}
\usepackage{soul}
\usepackage{url}
\usepackage[hidelinks]{hyperref}
\usepackage[utf8]{inputenc}
\usepackage[small]{caption}
\usepackage{graphicx}
\usepackage{amsmath}
\usepackage{amsthm}
\usepackage{booktabs}
\usepackage{algorithm}
\usepackage{algorithmic}
\usepackage[switch]{lineno}

\usepackage{graphicx}
\usepackage{amsfonts}
\usepackage{multirow}
\usepackage{colortbl}
\usepackage{makecell}
\usepackage{colortbl}
\usepackage{xcolor}
\usepackage{array}
\usepackage{mathrsfs}

\title{Improving Adversarial Robustness via Feature Pattern Consistency Constraint}

\author{
Jiacong Hu$^{1,3,4}$\and
Jingwen Ye$^{5}$\and
Zunlei Feng$^{2,3,4,}$\footnote{Corresponding author}\and
Jiazhen Yang$^{2}$\and
Shunyu Liu$^{1}$\and
Xiaotian Yu$^{1}$\and
Lingxiang Jia$^{1}$\And
Mingli Song$^{1,3,4}$
\affiliations
$^1$College of Computer Science and Technology, Zhejiang University\\
$^2$School of Software Technology, Zhejiang University\\
$^3$State Key Laboratory of Blockchain and Security, Zhejiang University\\
$^4$Hangzhou High-Tech Zone (Binjiang) Institute of Blockchain and Data Security\\
$^5$National University of Singapore\\
\emails
jiaconghu@zju.edu.cn,
jingweny@nus.edu.sg,
zunleifeng@zju.edu.cn,
yangjiazhen0915@outlook.com,
\{liushunyu,yuxiaotian,lingxiangjia,brooksong\}@zju.edu.cn
}

\begin{document}

\maketitle

\begin{abstract}
Convolutional Neural Networks (CNNs) are well-known for their vulnerability to adversarial attacks, posing significant security concerns. In response to these threats, various defense methods have emerged to bolster the model's robustness.
However, most existing methods either focus on learning from adversarial perturbations, leading to overfitting to the adversarial examples, or aim to eliminate such perturbations during inference, inevitably increasing computational burdens.
Conversely, clean training, which strengthens the model's robustness by relying solely on clean examples, can address the aforementioned issues.
In this paper, we align with this methodological stream and enhance its generalizability to unknown adversarial examples. This enhancement is achieved by scrutinizing the behavior of latent features within the network. 
Recognizing that a correct prediction relies on the correctness of the latent feature's pattern, we introduce a novel and effective Feature Pattern Consistency Constraint (FPCC) method to reinforce the latent feature's capacity to maintain the correct feature pattern. 
Specifically, we propose Spatial-wise Feature Modification and Channel-wise Feature Selection to enhance latent features. Subsequently, we employ the Pattern Consistency Loss to constrain the similarity between the feature pattern of the latent features and the correct feature pattern. Our experiments demonstrate that the FPCC method empowers latent features to uphold correct feature patterns even in the face of adversarial examples, resulting in inherent adversarial robustness surpassing state-of-the-art models.
\end{abstract}

\section{Introduction}
\label{sec:intro}

As deep learning continues to evolve, convolutional neural networks (CNNs) have gained widespread application across various domains, including facial recognition~\cite{luo2022memory}, autonomous driving~\cite{prakash2021multi}, and person identification~\cite{pu2023memorizing}.
However, despite their remarkable performance, CNNs exhibit a notable vulnerability to adversarial attacks.
Specifically, adversarial attacks generate samples imbued with minute perturbations that are imperceptible to humans but can mislead the model into making erroneous predictions~\cite{szegedy2013intriguing,goodfellow2014explaining}, posing a significant threat to security~\cite{wang2023detecting,liu2023transmission,ma2023llm}.

In the realm of defending against adversarial attacks, a plethora of methods has been proposed, with the majority concentrating on the utilization or elimination of adversarial perturbations generated by adversarial attacks. Adversarial training stands out as a widely acknowledged method for leveraging adversarial perturbations~\cite{madry2017towards,zhang2019theoretically} to enhance model robustness. This approach incorporates adversarial examples during model training, allowing the model to adapt to perturbations. However, a limitation of these methods is the potential overfitting of the trained model to specific types of attacks encountered during training~\cite{madry2017towards}, often leading to a decline in standard accuracy~\cite{laidlaw2020perceptual}.
On the other hand, for the elimination of adversarial perturbations, adversarial purification has gained significant traction recently~\cite{wang2022guided,nie2022diffusion}. This method employs additional generative models to remove adversarial perturbations from the adversarial examples. However, it does not inherently bolster the model's robustness and may contribute to increased computational costs during inference~\cite{croce2022evaluating}.
The limitations imposed by the aforementioned methods, which concentrate on adversarial perturbations, raise a crucial question: \emph{Can we break free from the paradigm of adversarial perturbations to enhance model robustness?}

\begin{figure*}[t!]
    \centering
    \includegraphics[scale=0.9]{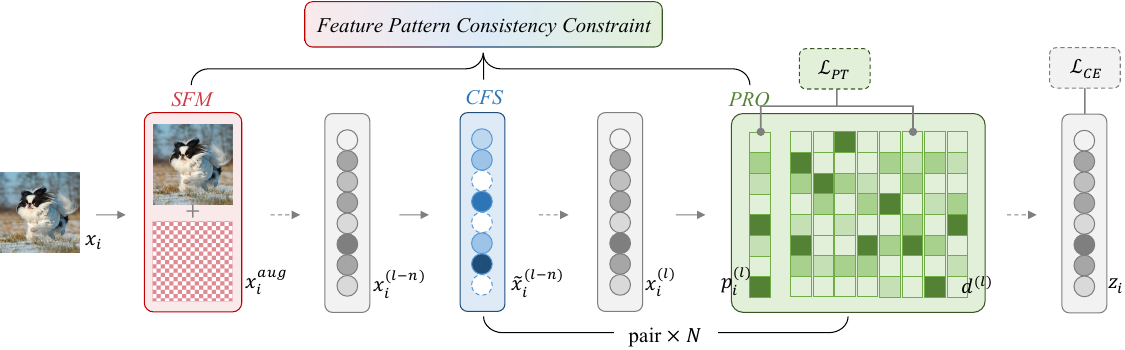}
    \caption{Feature Pattern Consistency Constraint training framework. Typically, `CFS' and `PRO' are configured in pairs within the network.}
    \label{fig:framework}
\end{figure*} 

Indeed, several methods have been proposed to enhance model robustness, relying solely on clean examples~\cite{mustafa2019adversarial,pang2019rethinking,li2021improving}. The majority of these methods aim to reduce intra-class distances, thereby increasing inter-class margins. However, the augmented margin is established within the feature space of clean examples, posing a challenge in adapting to adversarial perturbations and resulting in inferior robustness compared to adversarial training. Furthermore, the connection between the behavior of latent features and correct predictions in clean examples within these studies has not been thoroughly investigated.

In this paper, we redirect our focus to clean examples, delving into the behavior of latent features within the model. Specifically, through an analysis of feature behavior during correct predictions, we posit that correct predictions occur only when the \emph{feature pattern} is correct. This pattern is defined as the relative size among dimensions in the latent feature vector. Motivated by this insight, we propose a novel and effective Feature Pattern Consistency Constraint (FPCC) method to enhance the latent feature's capability in upholding the correct feature pattern.
Within FPCC, Spatial-wise Feature Modification and Channel-wise Feature Selection are introduced, approaching the problem from the perspective of adding noise and reducing features. This is done to improve robustness in scenarios with increasing interferential features and decreasing critical features when adversarial perturbations are added to clean examples. Subsequently, a Pattern-based Robustness Optimization is presented to constrain the similarity between the feature pattern of latent features and the correct feature pattern.
Experiments demonstrate that the proposed FPCC method enables latent features to uphold correct feature patterns even in the face of adversarial examples, leading to the model's inherent adversarial robustness surpassing state-of-the-art methods.

Our contributions can be summarized as follows:
\begin{itemize}
\item This paper introduces a novel and effective FPCC method, which comprises Spatial-wise Feature Modification and Channel-wise Feature Selection, along with the Pattern-based Robustness Optimization technique.

\item The proposed FPCC method is plug-and-play, seamlessly integratable into deep neural networks during training to augment the intrinsic robustness of the model, all without incurring any additional computational overhead during inference.

\item Extensive experiments demonstrate the effectiveness of the proposed FPCC method, showcasing superior adversarial robustness and inference speed compared to state-of-the-art methods.
\end{itemize}

\section{Related Work}
Adversarial defense methods can be categorized into two distinct types: static defense and adaptive test-time defense.

\subsection{Static Defense}
Within the realm of static defense, both the inputs and the parameters of the model remain constant during the inference process. One of the most effective defenses in this category is adversarial training~\cite{madry2017towards}, which involves training models with adversarial examples~\cite{zhang2019theoretically,cheng2020cat,gowal2020uncovering}. However, many adversarial training approaches can only defend against specific attacks they were trained with~\cite{madry2017towards} and often experience a significant accuracy drop on clean data~\cite{laidlaw2020perceptual}.

Conversely, some methods~\cite{li2021improving,mustafa2019adversarial,pang2019rethinking} focus on improving adversarial robustness solely by relying on clean examples to avoid overfitting to adversarial perturbations. Many of these methods aim to reduce inter-class distances, thereby increasing inter-class margins. For instance, Mustafa et al.~\cite{mustafa2019adversarial} introduced an approach to enhance robustness by compelling the features for each class to reside within a convex polytope that is maximally separated from the polytopes of other classes. However, the increased margin is based on the feature space of clean examples, thus achieving poor robustness when countering adversarial perturbations compared to adversarial training. Therefore, some margin-based methods~\cite{mustafa2019adversarial,pang2019rethinking} need to be combined with adversarial training to further improve robustness.

Different from the aforementioned methods, the proposed FPCC comprises strategies to address the situation of increasing interferential features and decreasing critical features when adversarial perturbations are added to clean examples, along with techniques that constrain the features to uphold the correct feature pattern.

\subsection{Adaptive Test-time Defense}
Adaptive test-time defense constitutes another pivotal category within adversarial defense methods, wherein the inputs or parameters of the model undergo dynamic alterations during the test phase. Adversarial purification~\cite{wang2022guided,nie2022diffusion,hill2020stochastic} is one of the most popular and effective methods in this category. It utilizes a generative model (e.g., GANs~\cite{goodfellow2014generative}, Diffusion models~\cite{ho2020denoising,song2020score,ma2023deepcache,li2023one}) to remove perturbations from adversarial examples. While these methods are plug-and-play and successfully defend against most attacks, they do incur significant additional computational costs during inference and do not inherently enhance the model's robustness~\cite{croce2022evaluating}.

Furthermore, some works~\cite{wang2021fighting,chen2021towards,kang2021stable,dong2022random,fu2021double} propose modifying parameters or activations during inference, essentially aiming to reduce the impact of perturbations during network prediction. However, these methods also incur computational overhead, and the improved robustness achieved through these approaches is limited, owing to the complexity of adversarial perturbation and the lack of diversity in training data.

In contrast to adaptive test-time defense methods, the proposed FPCC does not require integrating additional modules or dynamically adjusting defense strategies during inference. Therefore, it incurs no additional computational overhead.

\section{Method}
In this section, we conduct an analysis of feature behavior in the final layer of classification networks for correctly predicted samples, leading to the introduction of the concept of \emph{feature pattern}. We extend this concept to other layers of the network, suggesting that a network only achieves correct predictions when its latent features align with these correct feature patterns.
Furthermore, we propose the FPCC, which includes Spatial-wise Feature Modification and Channel-wise Feature Selection to address the situation of increasing interferential features and decreasing critical features when adversarial perturbations are added to clean examples. Additionally, FPCC incorporates Pattern-based Robustness Optimization to constrain the modified and selected features, ensuring the maintenance of the correct feature pattern.

\subsection{The Feature Pattern}
Given a classification network with $K(K\geq2)$ categories, the softmax function is typically used to calculate the probability of an input sample $x_i$ belonging to the $y_i$-th category, where $y_i \in \{1,2,\dots,K\}$ is the true label of $x_i$.
For the feature vector $z_i \in \mathbb{R}^K$ of $x_i$ at the final layer of the network,
the softmax function is defined as follows:
\begin{equation}
    \mathcal{S}({z_{i}[y_i]}) = \frac{\exp({z_{i}[y_i]})}{\sum_{k=1}^{K}\exp({z_{i}[k]})} \text{,}
\end{equation}
where $z_{i}[k]$ is the value in the $k$-th dimension of $z_i$.
Given that $y_i$ is the true label of $x_i$, the network is usually trained using cross-entropy loss to maximize the probability $\mathcal{S}({z_{i}[y_i]})$:
\begin{equation}
\mathcal{L}_{CE,i} = -log(\mathcal{S}({z_{i}[y_i]})) \text{.}
\end{equation}

Generally, the network's ability to correctly predict the label of $x_i$ is not solely based on the specific values of $z_i$, but rather on whether the value in the $y_i$-th dimension $z_{i}[y_i]$ is relatively larger than the values in other dimensions. We define this relative size among dimensions in the feature vector as the \emph{feature pattern}. In the network's final layer, each category has a distinct feature pattern, similar to the true label's one-hot vector. Correct predictions are made when the sample's feature pattern is correct, meaning it closely resembles the true category's feature pattern.
We propose that this concept of feature pattern can be generalized to other layers of the network.

\begin{figure}[t!]
    \centering
    \includegraphics[scale=0.9]{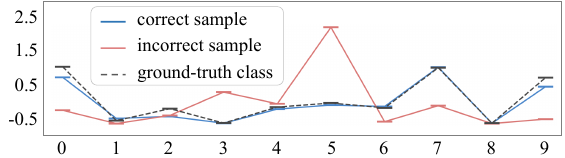}
    \caption{Feature patterns of \textcolor[RGB]{75, 135, 203}{a correctly predicted sample}, \textcolor[RGB]{218, 120, 118}{an incorrectly predicted sample}, and their corresponding \textcolor[RGB]{71, 71, 71}{ground-truth category}. The feature pattern of the ground-truth category is derived by averaging the feature patterns of the top 10 correctly predicted samples, identified based on the highest predicted probabilities.
    The horizontal axis denotes the feature dimensions, while the vertical axis represents the relative magnitude of the features. Both correct and incorrect samples are randomly selected from the 'dog' category of the CIFAR-10 dataset. To streamline the illustration, only the first $10$ dimensions of the penultimate layer (fully connected layer) of the VGG-16 network are displayed.}
    \label{fig:mt-fp}
\end{figure}

Next, we formalize the definition of the feature pattern, taking the $l$-th layer as an example and assuming it to be a fully connected layer (the case of convolutional layers will be discussed later). The feature pattern $p_i^{(l)}$ for a given sample $x_i$ at the $l$-th layer is articulated as the z-score normalization of the features $x_i^{(l)}$ at the same layer. This normalization procedure is employed to eliminate the feature scale, thereby upholding solely the relative sizes of the features across each dimension:
\begin{equation}
\label{eqn:fp}
\centering
\begin{aligned}
    p_i^{(l)} &= \frac{x_i^{(l)}-\overline{x}_i^{(l)}}{\sigma(x_i^{(l)}) + \epsilon} \text{,} \\
\end{aligned}
\end{equation}
where $\sigma(x_i^{(l)}) = \sqrt{\frac{1}{D^{(l)}}\sum_{d=1}^{D^{(l)}}{(x_{i}^{(l)}[d] - \overline{x}_i^{(l)}})}$, $\overline{x}_i^{(l)} = \frac{1}{D^{(l)}}\sum_{d=1}^{D^{(l)}}x_{i}^{(l)}[d]$, $D^{(l)}$ is the dimensionality of the features at the $l$-th layer, and $x_{i}^{(l)}[d]$ denotes the value of the $d$-th dimension within the feature $x_i^{(l)}$.

For a convolutional layer at the $l$-th layer, in contrast to fully connected layers
, the latent features are derived by averaging the output feature map:
\begin{equation}
x_i^{(l)}=\frac{1}{H^{(l)}}\frac{1}{W^{(l)}}\sum_{h=1}^{H^{(l)}}\sum_{w=1}^{W^{(l)}}f_i^{(l)} \text{,}
\end{equation}
where $f_i^{(l)} \in \mathbb{R}^{D^{(l)} \times H^{(l)} \times W^{(l)}}$ denotes the feature map of $x_i$ at the $l$-th layer, with $H^{(l)}$ and $W^{(l)}$ representing the height and width of the feature map at the $l$-th layer, respectively. Subsequently, the corresponding feature pattern $p_i^{(l)}$ can be determined using Eqn.~(\ref{eqn:fp}) as delineated above.

As depicted in Fig.~\ref{fig:mt-fp}, within an intermediate layer of the network, the feature pattern of incorrectly predicted samples deviates significantly from the feature pattern of the ground-truth category. In contrast, the feature pattern of correctly predicted samples closely mirrors the feature pattern of the ground-truth category. Consequently, akin to the final layer of the network, at the $l$-th layer, correct predictions hinge on the correctness of the feature pattern $p_i^{(l)}$ for $x_i$, i.e., its similarity to the feature pattern $d_{y_i}^{(l)}$ of the true category.
The similarity between the feature pattern of a sample $x_i$ and the ground-truth category $y_i$ can be quantified by the L1 distance between them:
\begin{equation}
    ||p_i^{(l)}-d_{y_i}^{(l)}||_1 \to 0 \Rightarrow y_i = \mathop{\arg\max}\limits_{k}\mathcal{S}({z_{i}[k]}) \text{,}
\end{equation}
where $d^{(l)} \in \mathbb{R}^{K \times D^{(l)}}$ represents the feature patterns of all categories at the $l$-th layer, and $d_{y_i}^{(l)}$ signifies the feature pattern of the $y_i$-th category.
Therefore, the enhancement of network robustness is contingent upon fortifying the capacity of features to uphold the correct feature patterns.

\subsection{Feature Pattern Consistency Constraint}
The introduced FPCC, designed to enhance the ability of features in upholding correct feature patterns, encompasses two strategies: Spatial-wise Feature Modification and Channel-wise Feature Selection. These strategies are devised to address the challenge posed by the increasing interferential features and decreasing critical features. Additionally, the FPCC incorporates a technique known as Pattern-based Robustness Optimization, which serves the purpose of constraining features to uphold the correct feature pattern.

\subsubsection{Spatial-wise Feature Modification (SFM)}
In order to address the challenge of increasing interferential features arising from the addition of adversarial perturbations to clean examples, we introduce Spatial-wise Feature Modification (SFM) to introduce controlled noise into the features.

Specifically, within SFM, a noise term $\delta$ is randomly generated within a budget of $\epsilon$ and subsequently added to the sample $x_i$:
\begin{equation}
\label{eqn:SFM}
\begin{aligned}
    x^{aug}_{i} = x_i + \delta \text{, }
    \delta \in (-\epsilon, +\epsilon) \text{.} 
\end{aligned}
\end{equation}

It is crucial to highlight that Eqn.(\ref{eqn:SFM}) differs from the generation of adversarial examples for adversarial training~\cite{madry2017towards} through adversarial attacks, as expressed by the following equation:
\begin{equation}
\begin{aligned}
    x_{i}^{adv} = x_i + \delta_i^{adv} \text{, }
    \delta_i^{adv} = \mathop{\arg\max}_{\|\delta_i^{adv}\| \leq \epsilon} \mathcal{L}_{CE}(x_i+\delta_i^{adv},y_i) \text{.}
\end{aligned}
\end{equation}
Our objective is not to adapt the model specifically to adversarial perturbations but rather to randomly generate noise, via SFM, to simulate the increase in interferential features. Subsequently, we aim to constrain these features to uphold correct feature patterns.
Additional insights from comparative distillation experiments pertaining to this aspect can be found in Section~\ref{sec:as-effect}.


\subsubsection{Channel-wise Feature Selection (CFS)}
To address the challenge of decreasing critical features resulting from the addition of adversarial perturbations to clean examples, we introduce Channel-wise Feature Selection (CFS). This approach involves the random selection of a subset of features, followed by the imposition of constraints to maintain the consistency of the feature pattern.
In CFS, a random subset of features is selected from a layer preceding the $l$-th layer (e.g., the $(l-n)$-th layer) utilizing a Bernoulli distribution:
\begin{equation}
\label{eqn:cfs}
\begin{aligned}
    {\widetilde{x}_{i}^{(l-n)}} = x_{i}^{(l-n)} \cdot b_{i}^{(l-n)} \text{, }
    b_{i}^{(l-n)} \sim B(\gamma) \text{,}
\end{aligned}
\end{equation}
where $b_{i}^{(l-n)} \in \mathbb{R}^{D^{(l-n)}}$ is drawn from a Bernoulli distribution with probability $\gamma$.
Subsequently, the selected features from the $(l-n)$-th layer, after undergoing $n$ layers of linear operations, are constrained to maintain the correct feature pattern at the $l$-th layer.

It is important to highlight that Eqn.~(\ref{eqn:cfs}) diverges from the concept of Dropout~\cite{srivastava2014dropout}, where a scaling factor of $\frac{1}{1-\gamma}$ is utilized to preserve the expected value of the features:
\begin{equation}
\begin{aligned}
    x_{i}^{(l-n)} = x_{i}^{(l-n)} \cdot b_{i}^{(l-n)} \times \frac{1}{1-\gamma} \text{, }
    b_{i}^{(l-n)} \sim B(\gamma) \text{.}
\end{aligned}
\end{equation}
The proposed FPCC exclusively constrains the selected original features $x_{i}^{(l-n)}$, rather than the scaled features $x_{i}^{(l-n)} \times \frac{1}{1-\gamma}$. Comprehensive ablation studies on this aspect are detailed in Section~\ref{sec:as-effect}.

\begin{algorithm}[!t]
\caption{Training Model with PRO} 
\label{alg:training} 
\begin{algorithmic}
    \REQUIRE{Training data $\{x_i\}$, ground truth labels $\{y_i\}$, initialized parameters $\theta$ in classification network, trainable parameters $\{d^{(l)}|l\in N\}$, hyperparameter $\lambda$, epochs $T$, the number of iteration $t \gets 0$.}
    \ENSURE{The parameters $\theta$.}
    \FOR {t to T}
    \STATE Compute the joint loss by $\mathcal{L} = \mathcal{L}_{CE} + \lambda\sum_{l \in N}\mathcal{L}_{PT}^{(l)}$.
    \STATE Compute the gradients $\frac{\partial\mathcal{L}}{\partial{x_i}}=\frac{\partial{\mathcal{L}_{CE}}}{\partial{x_i}} + \lambda\sum_{l \in N}\frac{\partial{\mathcal{L}_{PT}}}{\partial{x_i}}$
    \STATE Update the parameters $\theta$ by $\theta=\arg\min_\theta\mathcal{L}$.
    \STATE Update the parameters $d^{(l)}, \forall l \in N$. 
    \ENDFOR
\end{algorithmic} 
\end{algorithm}

\subsubsection{Pattern-based Robustness Optimization (PRO)}
\label{sec:pro}
To ensure that features, modified and selected by SFM and CFS, uphold correct feature patterns—specifically, ensuring the feature pattern of a sample aligns with the feature pattern of its ground-truth category—we propose Pattern-based Robustness Optimization (PRO). PRO constrains the L1 distance between the feature pattern of a sample at different layers (e.g., the $l$-th layer) and the feature pattern of its ground-truth category:
\begin{equation}
\label{eqn:fpc}
    \mathcal{L}_{PT, i}^{(l)} = ||p_i^{(l)}-d_{y_i}^{(l)}||_1 \text{.}
\end{equation}
It is crucial to emphasize that CFS and PRO are combined, as illustrated in Fig.~\ref{fig:framework}, and are incorporated into various layers of the network.
A comprehensive set of ablation studies pertaining to this aspect is presented in Section~\ref{sec:as-posqua}.

\begin{table*}[!t]
\small
\centering
\setlength{\tabcolsep}{3.2mm}{
\begin{tabular}{l|c|ccc|ccc|cc}
    \toprule
    & Clean & PGD  & AutoPGD & EOT-PGD & FGSM  & BIM & MIM & PIM & PIM++ \\
    \midrule
    \multicolumn{10}{l}{\cellcolor{green!10} WRN-28-10} \\
    \hline
    PGDAT &87.24    &61.38 / 50.71  &61.11 / 50.31  &60.16 / 50.51  &58.86  &50.70  &51.48    &19.36  &19.50 \\
    TRADES &84.60   &57.82 / 50.71  &57.50 / 50.30  &58.62 / 50.71  &59.57  &50.75  &51.75    &25.58  &23.20 \\
    AWP &84.09  &63.59 / 57.68  &63.44 / 57.35  &62.45 / 57.79  &62.92  &57.68  &58.08    &31.23  &30.75 \\
    MARGIN &85.80   &65.91 / 62.34  &65.76 / 62.10  &65.91 / 62.34  &67.10  &62.34  &62.82    &38.46  &37.29 \\
    SCORE &87.84    &67.04 / 63.75  &66.80 / 63.40  &65.04 / 63.25  &68.82  &63.69  &64.34    &38.26  &37.24 \\
    \emph{FPCC}(Ours) &\textbf{88.64}   &\textbf{67.28} / \textbf{65.58}    &\textbf{67.14} / \textbf{62.84}    &\textbf{66.90} / \textbf{65.72}    &\textbf{69.33} &\textbf{65.97} &\textbf{66.49} &\textbf{57.95} &\textbf{59.01} \\
    \hline
    RHS &90.45  &27.68 / 24.20  &26.96 / 22.78  &27.65 / 23.14  &43.65  &36.89  &38.03    &14.56  &14.36 \\
    RHS+AT &91.89   &55.13 / 46.43  &55.00 / 46.09  &61.14 / 46.43  &57.86  &49.56  &50.89    &22.47  &21.38 \\
    MMC &\textbf{92.70} &29.02 / 25.12  &28.46 / 24.97  &29.23 / 25.32  &44.70  &38.43  &39.95    &16.68  &16.17 \\
    MMC+AT &81.80   &60.34 / 55.03  &57.92 / 53.87  &61.35 / 55.43  &59.32  &52.86  &53.76    &24.32  &22.69 \\
    PROC &91.20 &29.64 / 26.60  &28.06 / 24.32  &29.66 / 26.63  &42.13  &33.65  &34.46    &15.03  &14.68 \\
    \emph{FPCC}(Ours) &88.64    &\textbf{67.28} / \textbf{65.58}    &\textbf{67.14} / \textbf{62.84}    &\textbf{66.90} / \textbf{65.72}    &\textbf{69.33} &\textbf{65.97} &\textbf{66.49} &\textbf{57.95} &\textbf{59.01} \\
    \midrule
    \multicolumn{10}{l}{\cellcolor{green!10} ResNet-50} \\
    \hline
    PGDAT &84.44    &61.00  / 51.80 &60.79 / 51.39  &61.20  / 50.95 &58.30  &51.76  &52.46    &22.15  &22.21 \\
    TRADES &84.05   &57.75 / 51.23  &57.52 / 50.73  &57.74 / 51.23  &59.63  &51.25  &52.32    &26.19  &23.29 \\
    AWP &79.32  &59.95 / 52.84  &59.81 / 52.75  &59.85 / 53.84  &56.22  &52.86  &53.18    &27.97  &28.65 \\
    MARGIN &83.02   &63.86 / 59.31  &63.63 / 59.14  &64.86 / 59.33  &63.15  &59.33  &59.67    &35.65  &34.99 \\
    SCORE &85.96    &66.34 / 61.63  &66.16 / 61.37  &\textbf{65.44} / 61.62 &66.42  &61.69  &62.16  &36.24  &35.4 \\
    \emph{FPCC}(Ours) &\textbf{87.65}   &\textbf{66.92} / \textbf{61.82}    &\textbf{66.53} / \textbf{61.39}    &65.23 / \textbf{62.72} &\textbf{67.21} &\textbf{61.92} &\textbf{62.22} &\textbf{52.29} &\textbf{54.52} \\
    \hline
    RHS &89.73  &24.95 / 22.36  &23.77 / 20.85  &24.76 / 21.39  &42.37  &37.54  &38.64  &15.12  &15.1 \\
    RHS+AT &91.26   &53.26 / 44.37  &52.81 / 44.24  &55.26 / 42.38  &58.42  &50.65  &51.36    &23.87  &22.54 \\
    MMC &\textbf{91.85} &25.68 / 22.46  &25.08 / 21.36  &25.66 / 22.45  &44.08  &39.08  &40.28    &17.26  &17.13 \\
    MMC+AT &82.46   &57.84 / 53.36  &54.43 / 51.22  &52.81 / 54.38  &59.21  &53.72  &54.65    &25.38  &24.33 \\
    PROC &90.48 &28.67 / 24.86  &24.78 / 22.67  &27.67 / 24.85  &41.86  &34.34  &35.14    &15.79  &15.74 \\
    \emph{FPCC}(Ours) &87.65    &\textbf{66.92} / \textbf{61.82}    &\textbf{66.53} / \textbf{61.39}    &\textbf{65.23} / \textbf{62.72}    &\textbf{67.21} &\textbf{61.92} &\textbf{62.22} &\textbf{52.29} &\textbf{54.52} \\
    \bottomrule
\end{tabular}
}
\caption{Standard accuracy (evaluated on clean data) and robust accuracy (against attacks) on CIFAR-10. Specifically, 'Clean' denotes the accuracy on clean data. 'Score1 / Score2' represents the accuracy against $\ell_2$ and $\ell_\infty$ attacks, respectively. For the remaining cases, 'Score' signifies the accuracy against $\ell_\infty$ attacks.}
\label{tab:exp-sota1}
\end{table*}

In PRO, the feature patterns of categories $d^{(l)}$ in the $l$-th layer are treated as learnable parameters and updated through gradient descent in each batch. Let $M$ denote the number of samples in a batch, and $N$ indicate the layers where feature patterns are constrained. The total loss in PRO is given as:
\begin{equation}
\begin{aligned}
    \mathcal{L} & = \mathcal{L}_{CE} + \lambda \sum_{l \in N} \mathcal{L}_{PT}^{(l)} 
    = \sum_{i=1}^{M}\mathcal{L}_{CE,i} + \lambda \sum_{i=1}^{M} \sum_{l \in N} \mathcal{L}_{PT,i}^{(l)} \\
    &= -\sum_{i=1}^{M} \log (\frac{\exp({z_{i}[y_i]})}{\sum_{k=1}^{K}\exp({z_{i}[k]})})  + \lambda \sum_{i=1}^{M} \sum_{l \in N} ||p_i^{(l)}-d_{y_i}^{(l)}||_1 \text{,}
\end{aligned}
\end{equation}
where $\lambda$ is a balancing factor for the two losses. The entire PRO process is end-to-end, as shown in Algorithm~\ref{alg:training}.

In summary, the proposed FPCC method incorporates SFM, CFS, PRO to bolster the network's inherent adversarial robustness. Notably, during the inference phase, the network obviates the need for the introduction of random noise, feature selection, or the constraint of sample feature patterns.

\section{Experiments}
\label{sec:experiments}

\subsection{Experimental Settings}
\paragraph{Datasets and Networks.}
We conducted experiments using two widely utilized datasets: CIFAR-10~\cite{krizhevsky2009learning} and CIFAR-100~\cite{krizhevsky2009learning}. Additionally, we evaluated the proposed method across three well-established classification networks: WRN-28-10~\cite{zagoruyko2016wide}, ResNet-50~\cite{he2016deep}, and VGG-16~\cite{simonyan2013deep}. As elaborated in Section~\ref{sec:pro}, CFS and PRO are combined and can be integrated into various layers of the networks. The specific configuration details of CFS and PRO are provided in the \emph{Supplementary Material}.

\paragraph{Adversarial Attacks.}
We assess the effectiveness of the proposed FPCC against a variety of formidable adaptive attacks, including the widely adopted Projected Gradient Descent (PGD)~\cite{madry2017towards} and its variant, AutoPGD~\cite{croce2020reliable}. To address the impact of randomization in model training, we also incorporate the Expectation Over Transformation (EOT) technique into PGD (referred to as EOT-PGD)~\cite{liu2018adv}, with EOT set to $10$ by default.
Additionally, we compare the performance of our method against single-step non-adaptive attacks and transferability-enhanced attacks. Specifically, we employ three common attacks: FGSM~\cite{goodfellow2014explaining}, BIM~\cite{kurakin2018adversarial}, and RFGSM~\cite{tramer2017ensemble}.
To demonstrate the broad applicability of our method beyond traditional full-pixel attacks, we evaluate its performance against
PIM~\cite{gao2020patch}, an iterative black-box patch attack method, and PIM++~\cite{gao2020patchpp}, a targeted patch attack approach.
Unless explicitly stated otherwise, the $\ell_{\infty}$ and $\ell_2$ attacks on models trained on the CIFAR-10 and CIFAR-100 datasets use a fixed budget of $\epsilon_\infty = 8/255$ and $\epsilon_2 = 0.5$, respectively.


\begin{table}[!t]
\small
\center
\setlength{\tabcolsep}{1.2mm}{
\begin{tabular}{l|c|ccc}
    \toprule
    & Clean & PGD  & AutoPGD & EOT-PGD \\
    \midrule
    \multicolumn{5}{l}{\cellcolor{green!10} VGG-16} \\
    \hline
    PGDAT &47.05	&32.93 / 30.28	&32.67 / 23.76	&32.92 / 31.27 \\
    TRADES &51.14	&32.70 / 31.22	&32.42 / 21.87	&32.58 / 31.12 \\
    AWP &49.49	&34.07 / 30.10	&33.89 / 30.89	&34.06 / 31.07 \\
    MARGIN &53.86	&36.88 / 34.52	&36.63 / 33.30	&36.87 / 34.42 \\
    SCORE &51.39	&35.51 / 34.22	&35.35 / 32.00 	&35.51 / 34.33 \\
    \emph{FPCC}(Ours) &\textbf{59.87}	&\textbf{39.74} / \textbf{38.69}	&\textbf{39.98} / \textbf{38.54}	&\textbf{39.69} / \textbf{38.04} \\
    \hline
    RHS &67.76	&11.69 / 10.10	&10.82 / 09.76	&12.67 / 11.10 \\
    RHS+AT &62.10	&34.27 / 29.88	&33.97 / 28.67	&34.37 / 29.84 \\
    MMC &\textbf{68.83}	&12.34 / 10.79	&11.36 / 10.08	&12.32 / 10.59 \\
    MMC+AT &52.67	&30.28 / 26.37	&29.78 / 25.46	&30.29 / 26.38 \\
    PROC &63.46	&13.36 / 12.68	&11.04 / 12.83	&12.38 / 11.79 \\
    \emph{FPCC}(Ours) &59.87	&\textbf{39.74} / \textbf{38.69}	&\textbf{39.98} / \textbf{38.54}	&\textbf{39.69} / \textbf{38.04} \\
    \bottomrule
\end{tabular}
}
\caption{
Standard accuracy and robust accuracy against $\ell_2$ and $\ell_\infty$ attack on CIFAR-100.
}
\label{tab:exp-sota2}
\end{table}

\subsection{Comparison with SOTA Methods}
Given that our method belongs to the realm of static defense, we selected multiple SOTA methods from this category for a fair comparison. These methods include adversarial training: PGDAT~\cite{madry2017towards}, TRADES~\cite{zhang2019theoretically}, AWP~\cite{wu2020adversarial}, MARGIN~\cite{gowal2020uncovering}, SCORE~\cite{pang2022robustness}, and the methods relying on clean examples: RHS~\cite{mustafa2019adversarial}, MMC~\cite{pang2019rethinking}, PROC~\cite{li2021improving}. The RHS and MMC can be combined with adversarial training in the original paper, denoted as RHS+AT and MMC+AT, respectively.
Additionally, we conducted comparisons between our method and adversarial purification methods. Due to space limitations, detailed results are provided in the \emph{Supplementary Material}.

Table~\ref{tab:exp-sota1} presents the accuracy against $\ell_2$ and $\ell_\infty$ attacks on CIFAR-10. The proposed FPCC method demonstrates higher standard accuracy compared to all adversarial training methods, and its robust accuracy surpasses methods relying solely on clean examples. For example, in WRN-28-10, the proposed FPCC achieves the highest accuracy of 65.58\% and 57.95\% against the PGD ($\ell_\infty$) attack and PIM attack, respectively. Similarly, in ResNet-50 against the PGD ($\ell_2$) attack, FPCC achieves a robust accuracy 41.97\% higher than RHS and surpasses the standard accuracy of PGDAT by 3.21\%.

Table~\ref{tab:exp-sota2} illustrates the accuracy against $\ell_2$ and $\ell_\infty$ attacks on CIFAR-100. The proposed FPCC achieves the highest accuracy. For instance, the standard accuracy of FPCC improves by 4.47\% compared to SCORE. Moreover, against AutoPGD ($\ell_2$) attack, FPCC improves accuracy by 28.94\% and 29.16\% compared to PROC and RHS, respectively.

\begin{table}[!t]
\small
\centering
\setlength{\tabcolsep}{2.2mm}{
\begin{tabular}{l|l|c|cc}
    \toprule
    & & Clean & White & Black  \\
    \midrule
    \multirow{2}{*}{\cellcolor{green!10}{VGG-16}} &
    Base &93.21	&00.00	&00.65 \\
    & \emph{FPCC} &89.42	&\textbf{74.37}	&\textbf{80.64} \\
    \hline
    \multirow{2}{*}{\cellcolor{green!10} ResNet-50} &
    Base &93.77	&00.00	&00.02 \\
    & \emph{FPCC} &87.65	&\textbf{61.82}	&\textbf{70.33} \\
    \hline
    \multirow{2}{*}{\cellcolor{green!10} WRN-28-10} &
    Base &94.54	&00.00	&00.03 \\
    & \emph{FPCC} &88.64	&\textbf{65.58}	&\textbf{71.98} \\
    \bottomrule
\end{tabular}
}
\caption{
Accuracy of models trained with and without FPCC against white-box and black-box attacks. Both attack types employed the PGD ($\ell_\infty$) attack. The experiments utilized the CIFAR-10 dataset.
}
\label{tab:ga-black}
\end{table}

\begin{table}[!t]
\small
\center
\begin{tabular}{l|ccc}
    \toprule
    \multirow{2}{*}{} & \multicolumn{3}{c}{Gradient ($\ell_2$ Norm)} \\
    & \cellcolor{green!10} VGG-16 & \cellcolor{green!10} ResNet-50 & \cellcolor{green!10} WRN-28-10 \\
    \hline
    Base & 0.38 & 0.55 & 0.46 \\
    \emph{FPCC} & 0.97 & 1.56 & 1.77 \\
    \bottomrule
\end{tabular}
\caption{
Average $\ell_2$ norm of gradients of the loss with respect to input images. The experiments were conducted on CIFAR-10.
}
\label{tab:ga-grad}
\end{table}

\subsection{Gradient Obfuscation}
Some studies~\cite{athalye2018obfuscated,carlini2019evaluating,tramer2020adaptive} have pointed out that the observed robustness improvements in many defense methods are due to gradient obfuscation, a phenomenon where the model's gradients are assumed to be hidden and unknown~\cite{athalye2018obfuscated}. To determine whether a defense method is based on gradient obfuscation, various works have suggested and employed several sanity checks, as outlined in~\cite{athalye2018obfuscated,carlini2019evaluating}, yet many methods have failed these tests. Key checks include:
1. Robustness against single-step attacks should be superior to that against iterative attacks.
2. Robustness against black-box attacks should be better than against white-box attacks.

We have demonstrated that our method passes these sanity checks. By comparing Tables~\ref{tab:exp-sota1} and ~\ref{tab:exp-sota2}, it is evident that our method exhibits better adversarial robustness against FGSM (a single-step attack) compared to PGD ($\ell_{\infty}$ and $\ell_2$, a iterative attack). Table~\ref{tab:ga-black} further shows that our method performs better under black-box attacks compared to white-box attacks.

Gradient vanishing is another form of gradient obfuscation. In Table~\ref{tab:ga-grad}, we assessed whether our proposed method suffers from gradient vanishing. It is observable that in our method, the $\ell_2$ norm of the gradients of the input image is not zero, indicating the absence of gradient vanishing.

\subsection{Ablation Studies}
\textbf{Effectiveness of SFM, CFS, and PRO.}
\label{sec:as-effect}
We first conducted ablation studies on SFM, CFS, and PRO, as shown in Table~\ref{tab:as-effect}. It is observed that the robust accuracy did not improve when the model was solely trained with SFM, CFS, or even PRO (indices `2', `3', and `4'). However, the proposed FPCC achieved a robust accuracy of 74.37\% (index `5'). The reason is that the latent features without modification (SFM) or selection (CFS) inherently possess the ability to uphold correct feature patterns, implying that there is no enhancement of features through the PRO.

\begin{figure}[!t]
    \centering
    \includegraphics[scale=0.85]{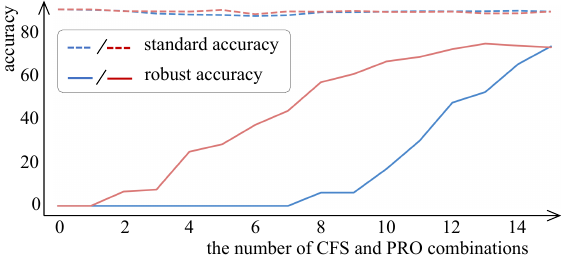}
    \caption{
    Impact of the positions and quantities of CFS and PRO on accuracy. The \textcolor[RGB]{75, 135, 203}{blue lines} and \textcolor[RGB]{218, 120, 118}{red lines} represent the cumulative insertion of CFS and PRO into the network, either from the shallow to the deep layer or from the deep to the shallow layer, respectively. Experiments were conducted on the VGG-16 using the CIFAR-10, and robust accuracy was measured against the PGD ($\ell_2$) attack.
    }
    \label{fig:as-posqua}
\end{figure}

\begin{table}[!t]
\center
\small
\setlength{\tabcolsep}{1.8mm}{
\begin{tabular}{l|c|l|cc}
    \toprule
    & Index & Modules & SA & RA  \\
    \midrule
    & 1 & Base & \textbf{93.21} & 0.00 \\
    \hline
    \cellcolor{green!10} Ours
    & 2 & + SFM & 92.08 & 0.11 \\
    & 3 & + CFS & 91.38 & 17.62 \\
    & 4 & + PRO & 92.64 & 2.82 \\
    & 5 & + SFM + CFS + PRO (\emph{FPCC}) & 89.42 & \textbf{74.37} \\
    \midrule
    \cellcolor{green!10}Adv.
    & 6 & + Adv. + CFS + PRO & 82.21 & 46.17 \\
    \midrule
    \cellcolor{green!10}Drop.
    & 7 & + SFM + Drop. + PRO & 87.63 & 3.07 \\
    \midrule
    \cellcolor{green!10}Cent.
    & 8 & + SFM + CFS + Cent. & 90.68 & 21.79 \\
    \bottomrule
\end{tabular}
}
\caption{Ablation studies for SFM, CFS, and PRO. `SA' and `RA' denote the standard accuracy evaluated on clean data and robust accuracy evaluated against the PGD ($\ell_2$) attack, respectively. `+Adv.' denotes adversarial examples produced via the FGSM attack. `Drop.' and `Cent.' correspond to Dropout and center loss, respectively. All experiments were conducted on the VGG-16 architecture using the CIFAR-10 dataset.}
\label{tab:as-effect}
\end{table}

In index `6', SFM in FPCC was replaced with adversarial examples, resulting in a robust accuracy of 46.17\%, while the standard accuracy dropped to 82.21\%. This decline is attributed to the model trained with adversarial examples becoming overfitted to adversarial perturbations.
In index `7', CFS in FPCC was replaced with Dropout. The robust accuracy of the model approached zero, as Dropout scales the selected features, causing PRO to constrain the scaled features rather than the original features.
Center loss~\cite{wen2016discriminative} is widely used in adversarial defense for reducing intra-class distances and thereby increasing inter-class margins~\cite{mustafa2019adversarial,pang2019rethinking}.
In index `8', PRO in FPCC was replaced with center loss, leading to a robust accuracy of 21.79\%. The improved robustness can be attributed to center loss promoting cohesiveness among intra-class features. However, it is less effective than PRO, as PRO constrains features to uphold correct feature patterns, aligning more closely with the observed behavior in correct predictions. Moreover, aligning features closer to a feature center does not necessarily align with the model's prediction logic.

\begin{figure}[!t]
    \centering
    \includegraphics[scale=0.85]{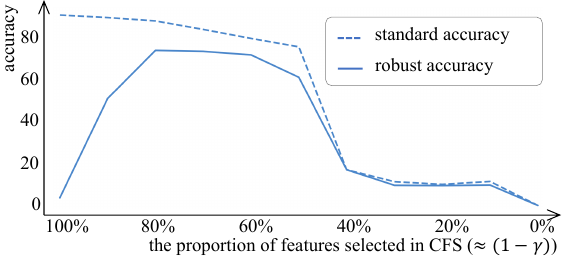}
    \caption{
    Impact of the proportion of features selected in CFS on accuracy. The variable $\gamma$ in Eqn.~\ref{eqn:cfs} is inversely proportional to the proportion of features selected; a larger $\gamma$ value corresponds to fewer features being selected. Experiments were conducted on the VGG-16 using the CIFAR-10. Robust accuracy was measured against the PGD ($\ell_2$) attack.
    }
    \label{fig:as-gamma}
\end{figure}

\textbf{Impact of the Positions and Quantities of CFS and PRO.}
\label{sec:as-posqua}
Fig.~\ref{fig:as-posqua} illustrates the impact of inserting varying numbers of CFS and PRO combinations into different layers of the network on accuracy. When CFS and PRO combinations are cumulatively inserted into the network from the shallow to the deep layer (blue line), it is observed that the robust accuracy of the model begins to rise continuously after the number of CFS and PRO combinations or the layer of the network reaches $7$. This suggests that merely adding CFS and PRO to the shallower layers of the network has little to no effect on enhancing model robustness.
Conversely, when CFS and PRO combinations are cumulatively inserted into the network from the deep to the shallow layer (red line), the robust accuracy of the model starts to increase from the initial stage. However, as more CFS and PRO combinations are inserted into the shallower layers, the robust accuracy of the model tends to decrease. This indicates that inserting CFS and PRO into the deeper layers of the network is more effective than inserting them into the shallower layers.

\textbf{Impact of the Proportion of Features Selected in CFS.}
Fig.~\ref{fig:as-gamma} illustrates the impact of the proportion of features selected in CFS on accuracy. As observed, when the proportion of selected features decreases from 100\% to 80\%, the robust accuracy of the model initially increases to its peak, while the standard accuracy experiences a slight decline. This is because when 100\% of the features are selected, all of them inherently possess the ability to uphold correct feature patterns, implying that there is no enhancement of features.
Furthermore, when the proportion of selected features falls below 50\%, both the robust and standard accuracies of the model decline sharply. This decrease can be attributed to the increased difficulty in upholding correct feature patterns with a smaller selection of features.

\section{Conclusion}
This paper diverges from most adversarial defense methods that focus on perturbations, directing attention instead to the clean examples. We introduce a novel and effective Feature Pattern Consistency Constraint (FPCC) method to strengthen the ability of features to uphold correct feature patterns. Experimental validation shows that the reinforced features can uphold correct patterns even when confronted with adversarial disturbances, enhancing the network's intrinsic adversarial robustness beyond current SOTA methods.
However, our approach has only been experimentally validated and lacks theoretical backing. Additionally, recent methods employing additional modules for adversarial purification, though not enhancing the model's inherent robustness, effectively remove perturbations from adversarial examples and surpass our method in terms of robust accuracy. Hence, there is significant room for research in enhancing model adversarial robustness from the perspective of clean training, which will be the focus of our future work.

\clearpage

\section*{Acknowledgments}
This work is supported by National Natural Science Foundation of China (U20B2066), Zhejiang Province High-Level Talents Special Support Program ``Leading Talent of Technological Innovation of Ten-Thousands Talents Program'' (No. 2022R52046), Zhejiang Province ``LingYan'' Research and Development Plan Project (No. 2024C01114).

\bibliographystyle{named}
\bibliography{ijcai24}

\end{document}